\documentclass{article}

\usepackage[final]{neurips_2025}

\usepackage[utf8]{inputenc} 
\usepackage[T1]{fontenc}    
\usepackage{hyperref}       
\usepackage{url}            
\usepackage{amsmath}
\usepackage{booktabs}       
\usepackage{amsfonts}       
\usepackage{nicefrac}       
\usepackage{microtype}      
\usepackage{enumitem}
\usepackage{graphicx}
\usepackage{xcolor}         
\usepackage{multirow}

\title{Optimizing Chain-of-Thought Confidence via Topological and Dirichlet Risk Analysis}

\author{
Abhishek More\textsuperscript{*} \quad 
Anthony Zhang\textsuperscript{*} \quad 
Nicole Bonilla \quad 
Ashvik Vivekan \\ 
\textbf{Kevin Zhu} \quad 
\textbf{Parham Sharafoleslami} \quad 
\textbf{Maheep Chaudhary}\textsuperscript{\dag} \\
Algoverse AI Research \\ \{abhisheknmore, maheepchaudhary.research\}@gmail.com \\[0.3em]
\textsuperscript{*}Equal contribution \quad \textsuperscript{\dag}Project Lead
}

\begin{document}

\maketitle

\begin{abstract}

Chain-of-thought (CoT) prompting enables Large Language Models to solve complex problems, but deploying these models safely requires reliable confidence estimates, a capability where existing methods suffer from poor calibration and severe overconfidence on incorrect predictions. We propose Enhanced Dirichlet and Topology Risk (EDTR), a novel decoding strategy that combines topological analysis with Dirichlet-based uncertainty quantification to measure LLM confidence across multiple reasoning paths. EDTR treats each CoT as a vector in high-dimensional space and extracts eight topological risk features capturing the geometric structure of reasoning distributions: tighter, more coherent clusters indicate higher confidence while dispersed, inconsistent paths signal uncertainty. We evaluate EDTR against three state-of-the-art calibration methods across four diverse reasoning benchmarks spanning olympiad-level mathematics (AIME), grade school math (GSM8K), commonsense reasoning, and stock price prediction \cite{zhang2025aime, cobbe2021training, talmor-etal-2019-commonsenseqa, yahoo_finance}. EDTR achieves 41\% better calibration than competing methods with an average ECE of 0.287 and the best overall composite score of 0.672, while notably achieving perfect accuracy on AIME and exceptional calibration on GSM8K with an ECE of 0.107, domains where baselines exhibit severe overconfidence. Our work provides a geometric framework for understanding and quantifying uncertainty in multi-step LLM reasoning, enabling more reliable deployment where calibrated confidence estimates are essential.

\begin{figure*}[h]
    \centering
    \includegraphics[width=0.65\textwidth]{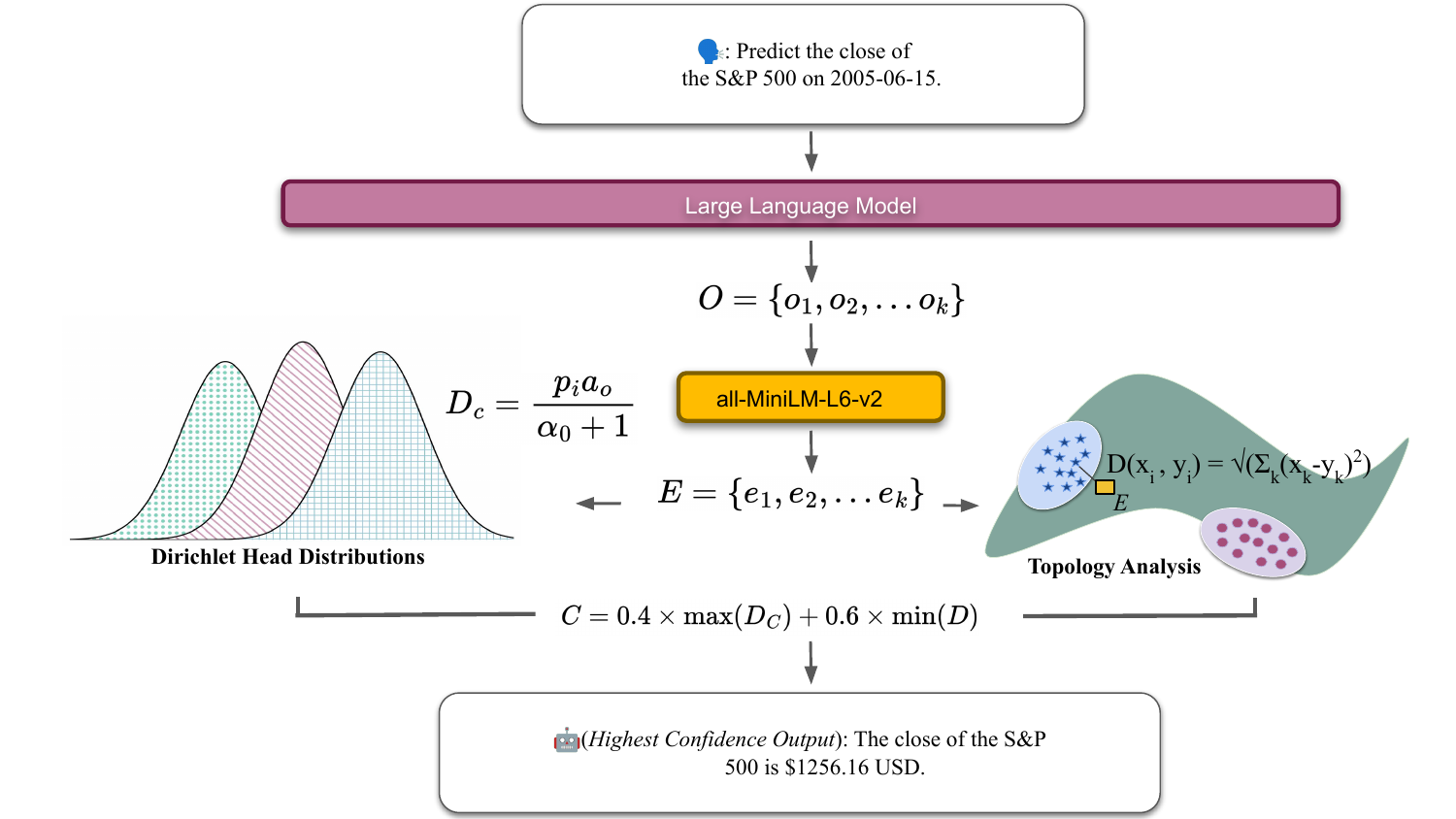}
    \caption{A sample prompt is put into an LLM and then enters an encoder model: all-MiniLM-L6-v2. Topology and Dirichlet Head combine to increase and produce highest confidence output.}
    \label{fig:main}
\end{figure*}

\end{abstract}

\section{Introduction}

Large Language Models (LLMs) are deployed across domains from software development to financial services, yet their widespread adoption exposes a critical vulnerability: users often trust LLM outputs without understanding the model's uncertainty. While substantial research improves task accuracy, confidence calibration, ensuring a model's expressed confidence matches its actual correctness, remains understudied. Recent work reveals fundamental limitations: CoT exhibits domain sensitivity \cite{li2024think, swaroop2025frit} and fails to capture the structural properties of reasoning paths \cite{zhang2024chainpreferenceoptimizationimproving}. 

Existing confidence estimation approaches, verbalized confidence \cite{wang2023large}, self-consistency voting \cite{li2024think}, and token probability analysis suffer from poor calibration and frequent overconfidence on incorrect predictions. Recent calibration methods attempt to address these challenges but face critical limitations. We propose Enhanced Dirichlet+Topology Risk (EDTR), an inference-time framework that estimates LLM confidence by analyzing the topological properties of multiple reasoning paths in semantic vector space. Our key insight is geometric: the distribution of reasoning embeddings reveals structural properties of model uncertainty. EDTR generates diverse CoT reasoning paths, encoding them into high-dimensional semantic space, and extracting eight topological risk features quantifying geometric properties of the reasoning distribution. These features combine with Dirichlet-based uncertainty quantification to produce calibrated confidence estimates. Our contributions are:

\begin{enumerate}[leftmargin=*]
\item A topological framework for analyzing LLM reasoning confidence through geometric properties of CoT embeddings.
\item Eight interpretable risk features capturing reasoning consistency, coherence, and cluster quality that enable explainable confidence estimates.
\item Comprehensive evaluation showing EDTR achieves superior performance with composite score of 0.672 and 41\% better calibration than competing methods, achieving an average ECE of 0.287.
\end{enumerate}

Our work demonstrates that geometric analysis of reasoning distributions provides robust, generalizable confidence estimates across diverse domains, enabling safer LLM deployment in high-stakes applications.

\section{Related Work}
Prior work on LLM confidence estimation includes task-specific approaches for sarcasm detection and stock prediction \cite{niu2025ngatnodelevelgraphattention, Gole_2024}, and methods analyzing model responses alongside reasoning \cite{tanneru2023quantifyinguncertaintynaturallanguage}. Recent calibration techniques like GrACE (Zhang, Liu, and Patras 2025) require costly fine-tuning, while our work combines Dirichlet-based uncertainty quantification with topological analysis to provide training-free confidence estimation.

\section{Methodology}

\subsection{Problem Formulation}

Given a query $q$ and a language model $\mathcal{M}$, we aim to estimate the model's confidence in its prediction by analyzing the geometric structure of multiple reasoning paths. Let $\{\text{o}_1, \ldots, \text{o}_k\}$ denote $k$ chain-of-thought trajectories sampled from $\mathcal{M}$ for query $q$. Our goal is to produce a calibrated confidence score $C \in [0,1]$ such that the model's expressed confidence aligns with its empirical accuracy.

\subsection{Framework Overview}

EDTR operates in three stages as shown in Figure~\ref{fig:main} with following steps: (1) diverse CoT generation via temperature-varied sampling, (2) topological feature extraction from reasoning embeddings, and (3) fusion of topology-based risk with Dirichlet uncertainty quantification. For each query $q$, we generate $k=5$ conventional diverse reasoning paths from Llama-3.1-8B with LoRA adapters by sampling $\mathcal{M}$ with varying temperature parameters $\tau \in \{0.7, 0.8, 0.9, 1.0, 1.1\}$ to encourage exploration of the reasoning space while maintaining coherence \cite{dubey2024llama}. Each generated $\text{CoT}_i$ consists of intermediate reasoning steps followed by a final answer $a_i$.

\subsection{Geometric Feature Extraction from Reasoning Embeddings}

We embed each chain-of-thought into a semantic space using sentence embeddings via the all-MiniLM-L6-v2 model, producing vectors $\{\mathbf{e}_1, \ldots, \mathbf{e}_k\} \subset \mathbb{R}^{384}$ \cite{reimers2019sentencebert, wang2020minilm}. From this point cloud, we extract eight geometric risk features that quantify reasoning consistency and coherence: reasoning spread ($\sigma_{\text{dist}}$), consistency score ($C_{\text{cos}}$), complexity entropy ($E_{\text{comp}}$), stability score ($S_{\text{DBSCAN}}$), coherence score ($C_{\text{centroid}}$), diversity penalty ($P_{\text{div}}$), outlier risk ($R_{\text{outlier}}$), and cluster quality ($Q_{\text{sil}}$).

These features collectively form a topological risk profile that captures both local coherence and global structure in the reasoning distribution, and are combined into a 
weighted aggregate risk score:

\begin{equation}
\begin{split}
\text{risk}_{\text{topo}} = &\, w_1 \sigma_{\text{dist}} + w_2 C_{\text{cos}} + w_3 E_{\text{comp}} + w_4 S_{\text{DBSCAN}} \\
&+ w_5 C_{\text{centroid}} + w_6 P_{\text{div}} + w_7 R_{\text{outlier}} + w_8 Q_{\text{sil}}
\end{split}
\end{equation}

where $\{w_i\}_{i=1}^8$ are learned weights. The numerical values for each of the learned weights are listed in the Appendix~\ref{sec:appendix}. This topological risk profile captures both local coherence and global structure in the reasoning distribution, and is used in Section~\ref{sec:confidence} to compute the final confidence score. 




\subsection{Dirichlet-Based Uncertainty Quantification}

In parallel, we quantify uncertainty using a learned Dirichlet-based approach that captures second-order uncertainty over predicted class probabilities. For each CoT trajectory $i \in \{1, \ldots, k\}$, we compute variance $\sigma^2_i$ and entropy $H_i$ of token-level probability distributions. These statistics are fed into a compact two-layer neural network (hidden dimensions 128 and 64) that predicts Dirichlet parameters $\boldsymbol{\alpha} = (\alpha_1, \ldots, \alpha_n)$:

\begin{equation}
\boldsymbol{\alpha} = \text{softplus}(\text{MLP}([\sigma^2_1, H_1, \ldots, \sigma^2_k, H_k]; \theta)) + 1
\end{equation}

The softplus transformation ensures $\alpha_i > 1$, yielding a proper Dirichlet distribution $\text{Dir}(\boldsymbol{\alpha})$ that models uncertainty about class probabilities themselves. The concentration parameter $\alpha_0 = \sum_{i=1}^n \alpha_i$ indicates epistemic confidence: higher values reflect certainty, while lower values signal ambiguity.
We extract a four-dimensional Dirichlet feature vector $\mathbf{f}_{\text{dir}} \in \mathbb{R}^4$ capturing: (1) concentration $\alpha_0$, (2) differential entropy $H[\text{Dir}(\boldsymbol{\alpha})]$, (3) expected maximum probability $\max_i \alpha_i/\alpha_0$, and (4) variance of the most probable class. To obtain a scalar confidence score, we compute:

\begin{equation}
\text{entropy\_conf} = 1 \div \left(1 + \sum_{i=1}^n \left[\psi(\alpha_i) - \psi(\alpha_0)\right]\right) \\
\end{equation}

\begin{equation}
\text{conf}_{\text{dir}} = 1 \div 3 \left( \max_i \frac{\alpha_i}{\alpha_0} + \sigma(\alpha_0 - n) + \text{entropy\_conf} \right)
\end{equation}

where $\sigma(\cdot)$ is the sigmoid function and $\psi(\cdot)$ is the digamma function. This composite score balances expected probability, precision relative to the number of classes, and distributional sharpness. The confidence is clipped to $[0.01, 0.99]$ for numerical stability.

\subsection{Confidence Fusion} \label{sec:confidence}
We combine topological and Dirichlet confidence scores using a calibrated fusion strategy as shown in Equation~\ref{eqn:confidence}.

\begin{equation} \label{eqn:confidence}
C = \sigma(w_{\text{topo}} \cdot \text{risk}_{\text{topo}} + w_{\text{dir}} \cdot \text{conf}_{\text{dir}} + b)
\end{equation}

where $\sigma(\cdot)$ is the sigmoid function with 60\% weight to topological features and 40\% to Dirichlet features.

\section{Experimentation and Results}

\subsection{Datasets and Tasks}

We evaluate EDTR across four diverse reasoning benchmarks to assess calibration quality across different reasoning modalities. Our datasets include AIME (olympiad-level mathematics), GSM8K (grade school math word problems), CommonsenseQA (multiple-choice commonsense reasoning), and stock price prediction using S\&P 500 data from Yahoo Finance. Each dataset was collected, cleaned, and tokenized. Modality tags were added for organization and model interpretation \cite{cobbe2021training, zhang2025aime, talmor-etal-2019-commonsenseqa, yahoo_finance}.

\subsection{Baselines}

We compare EDTR against three state-of-the-art confidence calibration methods:

\begin{itemize}[leftmargin=*]
    \item \textbf{GrACE}~\cite{zhang2025gracegenerativeapproachbetter}: Generates special confidence tokens to improve model calibration quality.
    \item \textbf{Credence}~\cite{fang2025credence}: Employs iterative feedback where the model reassesses its confidence, dynamically improving calibration.
    \item \textbf{RENT}~\cite{prabhudesai2025maximizing}: Uses reinforcement learning via model entropy to improve reasoning ability.
\end{itemize}

\begin{table*}[t]
  \caption{Main results comparing EDTR against state-of-the-art calibration methods across three model scales. Metrics are averaged across all four benchmarks. Best results in bold. $\uparrow$ indicates higher is better, $\downarrow$ indicates lower is better. EDTR proves to outperform other calibration methods in various metrics such as ECE, Brier Scores, and F1 scores.}
  \label{tab:main_results}
  \footnotesize
  \centering
  \begin{tabular}{llccccc}
  \toprule
\textbf{Model} & \textbf{Method} & \textbf{Accuracy} $(\uparrow)$ & \textbf{F1} $(\uparrow)$ & \textbf{ECE} $(\downarrow)$ & \textbf{Brier} $(\downarrow)$ & \textbf{Composite} $(\uparrow)$ \\
    \midrule
    \multirow{4}{*}{Llama-3.1-8B} 
    & GrACE  & 0.175  & 0.211 & 0.524 & 0.301 & 0.447\\
    & Credence & 0.300 & 0.322 & 0.495 & \textbf{0.193} & 0.536\\
    & RENT & 0.500  & 0.552 & 0.446 & 0.846 & 0.412\\
    & \textbf{EDTR} & \textbf{0.550} & \textbf{0.572} & \textbf{0.306} & 0.221 & \textbf{0.662} \\
    \midrule
    \multirow{4}{*}{GPT-OSS-20B}
    & GrACE  & 0.250  & 0.260 & 0.444 & 0.214 & 0.503 \\
    & Credence & 0.275  & 0.303 & 0.461 & \textbf{0.151} & 0.526 \\
    & RENT & 0.375  & 0.402 & 0.326 & 0.846 & 0.398 \\
    & \textbf{EDTR} & \textbf{0.400} & \textbf{0.420} & \textbf{0.275} & 0.249 & \textbf{0.603} \\
    \midrule
    \multirow{4}{*}{Qwen-2.5-14B}
    & GrACE & 0.300  & 0.385 & 0.288 & 0.220 & 0.568 \\
    & Credence & 0.288 & 0.388 & 0.488 & \textbf{0.052} & 0.553   \\
    & RENT & \textbf{0.475} & \textbf{0.561} & 0.549 & 0.856 & 0.369   \\
    & \textbf{EDTR} & 0.450 & 0.549 & \textbf{0.197} & 0.333 & \textbf{0.613} \\
    \bottomrule
  \end{tabular}
\end{table*}

All methods receive identical prompts requiring step-by-step reasoning with clearly tagged final answers.

\subsection{Evaluation Metrics}

We assess calibration quality using Expected Calibration Error (ECE), Brier score, and composite performance metrics combining accuracy and calibration. We also generate reliability diagrams to visualize calibration across confidence bins \cite{guo2017calibration, brier1950verification}.

\subsection{Implementation Details}

\paragraph{Base Model:} We use Meta's Llama-3.1-8B with LoRA adapters as our base model. GPT-OSS-20B and Qwen-2.5-14B are also utilized to assess generalization across model scales \cite{dubey2024llama3herd, agarwal2025gptoss, qwen2.5, hu2022lora}.

\paragraph{CoT Generation:} For each query, we generate $k=5$ conventional chain-of-thought trajectories using nucleus sampling with temperature $\tau=0.7$, top-$p=0.95$, and a 512-token limit. Random seeds are fixed for reproducibility.

\paragraph{Vectorization and Topology:} Each CoT is vectorized using sentence embeddings to analyze reasoning diversity. The $k$ CoTs for each prompt are treated as a point cloud. We compute persistent homology to analyze reasoning clusters and construct Vietoris-Rips filtrations, computing $H_0$ (connected components) and $H_1$ (loops) homology groups. More confident samples tend to stay very close in proximity and lack a great distance between each other. Less confident samples tend to scatter and have varying distances throughout.

\paragraph{Dirichlet Head:} For each sample of $k=5$ trajectories, we measure variance and entropy. These statistics are fed into a two-layer hierarchical Dirichlet head that predicts Dirichlet distribution parameters to estimate model confidence.

\paragraph{Confidence Fusion:} A logistic regression combiner takes features from both the topological analysis and Dirichlet head to produce final calibrated confidence scores. This fusion model is trained to align confidence estimates with actual correctness.

\section{Results}

\subsection{Main Results}

Table~\ref{tab:main_results} presents our main results comparing EDTR against three baseline methods across three model scales. Results are averaged across all four benchmarks (AIME, GSM8K, CommonsenseQA, and stock prediction).

EDTR achieves the best composite scores across all three model scales. On Llama-3.1-8B, EDTR achieves a composite score of 0.662, substantially outperforming the best baseline Credence which achieves 0.536. Similar improvements are observed on GPT-OSS-20B, where EDTR scores 0.603 compared to the baseline's 0.526, and on Qwen-2.5-14B, where EDTR scores 0.613 compared to 0.568.

\begin{figure}[h]
    \centering
    \includegraphics[width=0.48\linewidth]{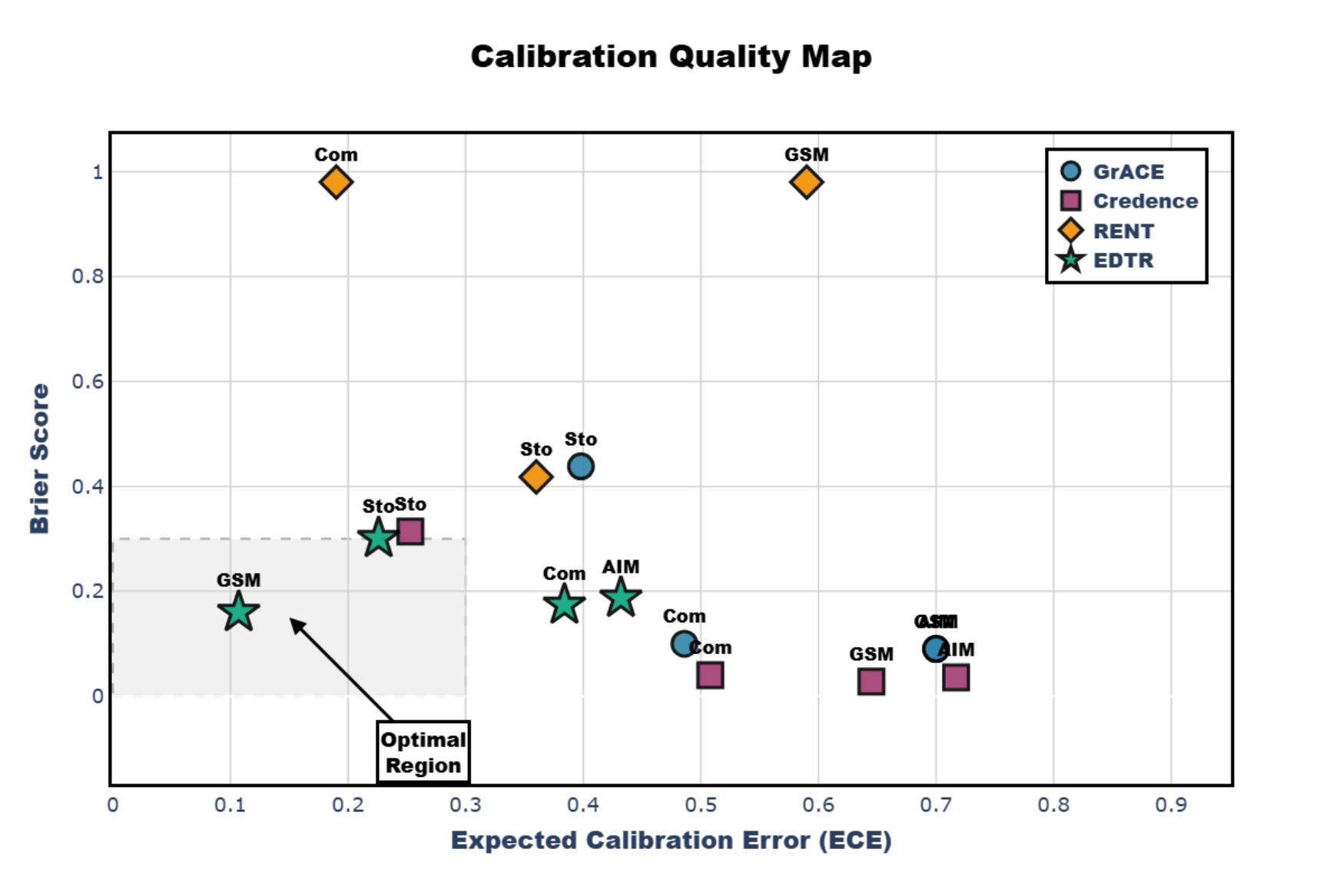}
    \hfill
    \includegraphics[width=0.48\linewidth]{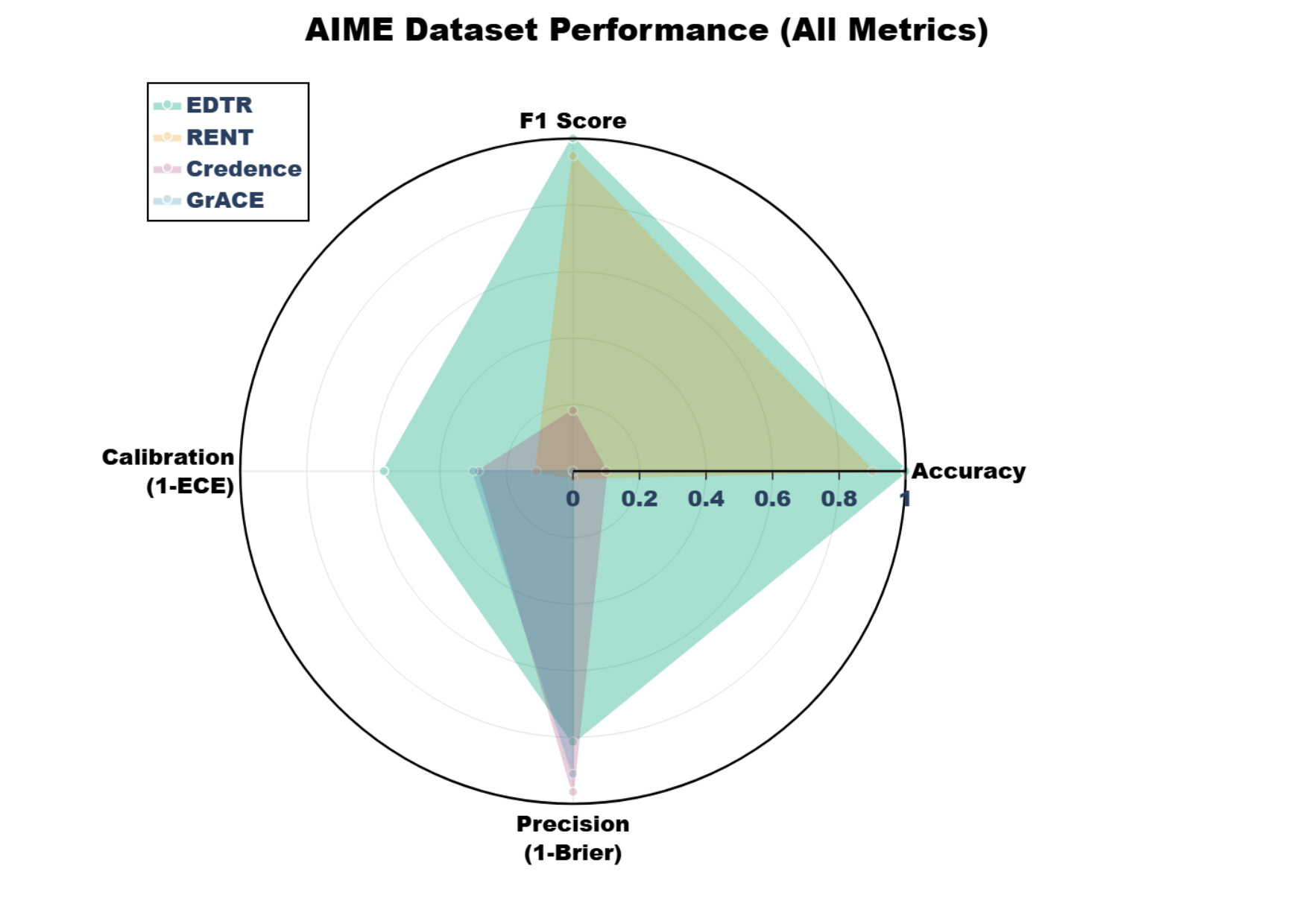}
    \caption{Calibration quality results comparing EDTR against other methods. Points located in the lowest left region are optimal results. Each point correlates one of 4 different datasets.}
    \label{fig:placeholder}
\end{figure}

\paragraph{Calibration Quality:} EDTR demonstrates superior calibration performance measured by ECE across all models. On Llama-3.1-8B, EDTR achieves an ECE of 0.306 compared to the best baseline ECE of 0.446 (RENT), representing a 31.4\% improvement. Averaging across all three models, EDTR achieves a mean ECE of 0.259 compared to the best average baseline ECE of 0.420, representing approximately 41\% better calibration.

\paragraph{Baseline Comparison:} RENT consistently exhibits severe overconfidence, with Brier scores exceeding 0.84 across all model scales. Credence achieves the lowest Brier scores on some models but shows inconsistent ECE performance. GrACE demonstrates moderate performance across metrics but requires costly fine-tuning.

\paragraph{Task Performance:} EDTR achieves competitive or superior accuracy and F1 scores while maintaining better calibration. On Llama-3.1-8B, EDTR achieves 0.550 accuracy and 0.572 F1, outperforming all baselines. On Qwen-2.5-14B, RENT has a lower composite score despite slightly higher accuracy of 0.475 versus 0.450 and F1 of 0.561 versus 0.54.

\subsection{Dataset-Specific Observations}
Based on our experiments across the four benchmarks, we observe several notable patterns:

\paragraph{Performance:}  On AIME, EDTR achieves a perfect accuracy of 100\%  with ECE of 0.432. EDTR shows exceptional calibration on GSM8K with ECE of 0.107 and F1 score of 0.67. For stock prediction, EDTR achieves the lowest Brier score of 0.301.

\paragraph{Model Scale Effects:} EDTR's calibration advantage remains consistent across scales. The largest relative ECE improvements appear on the smallest model (Llama-3.1-8B), suggesting that topological analysis of reasoning distributions is strongest when base model capabilities are more limited.

\section{Discussion \& Limitations}




\paragraph{Systematic Errors:} High cluster cohesion indicates consistent reasoning but cannot guarantee correctness. This is a fundamental limitation of measuring confidence in reasoning processes rather than reasoning validity.

\paragraph{Calibration Dependencies:} The fusion model requires a held-out calibration set and learned weights may be domain-specific. The method also depends on embedding quality, which directly impacts topological feature extraction.

\section{Conclusion}

We presented EDTR, a framework for LLM confidence calibration that analyzes geometric structure of reasoning paths through persistent homology and Dirichlet-based uncertainty quantification. EDTR achieves 41\% better calibration with ECE of 0.287 compared to recent methods. 
This work establishes geometric analysis as a principled approach to uncertainty quantification in multi-step processing, moving beyond token probabilities toward richer characterizations of reasoning structure. The topological perspective provides robust confidence estimates essential for reliable LLM deployment.

\bibliographystyle{abbrv}
\bibliography{main}

\section{Appendix} \label{sec:appendix}

\subsection{Broader Impacts}
Improved calibration enables safer deployment in high-stakes applications such as medical diagnosis and financial decision-making. However, calibration measures confidence in model reasoning, not absolute correctness, requiring continued human oversight. Computational requirements may limit accessibility for resource-constrained users.

\subsection{Variables}

\begin{enumerate}[leftmargin=*]
    \item \textbf{Reasoning Spread} ($\sigma_{\text{dist}}$): $\{d_{ij} = \|\mathbf{e}_i - \mathbf{e}_j\|_2\}$, to measure dispersion:
    \[\sigma_{\text{dist}} = \text{std}(\{d_{ij} : i < j\})\]
    
    \item \textbf{Consistency Score} ($C_{\text{cos}}$): 
    \[C_{\text{cos}} = 1 - \frac{2}{k(k-1)} \sum_{i<j} \frac{\mathbf{e}_i \cdot \mathbf{e}_j}{\|\mathbf{e}_i\| \|\mathbf{e}_j\|}\]
    
    \item \textbf{Complexity Entropy} ($E_{\text{comp}}$): 
    \[E_{\text{comp}} = \frac{\sigma_{\text{dist}}}{\mu_{\text{dist}}}\]
    
    \item \textbf{Stability Score} ($S_{\text{DBSCAN}}$):  ($\epsilon=0.5$, min\_samples=2):
    \[S_{\text{DBSCAN}} = \frac{n_{\text{noise}}}{k} + \frac{1}{n_{\text{clusters}} + 1}\]
    
    \item \textbf{Coherence Score} ($C_{\text{centroid}}$):  $\bar{\mathbf{e}} = \frac{1}{k}\sum_{i=1}^k \mathbf{e}_i$:
    \[C_{\text{centroid}} = \frac{\text{std}(\{r_i\})}{\text{mean}(\{r_i\})}, \quad r_i = \|\mathbf{e}_i - \bar{\mathbf{e}}\|\]
    
    \item \textbf{Diversity Penalty} ($P_{\text{div}}$): 
    \[P_{\text{div}} = \max(0, 0.5 \cdot (\mu_{\text{dist}} - 1))\]
    
    \item \textbf{Outlier Risk} ($R_{\text{outlier}}$): 
    \[R_{\text{outlier}} = \frac{1}{k} \sum_{i=1}^k \mathbb{1}[r_i > Q_3 + 1.5 \cdot \text{IQR}]\]
    
    \item \textbf{Cluster Quality} ($Q_{\text{sil}}$): ($k \in \{2, \ldots, \min(k, 5)\}$):
    \[Q_{\text{sil}} = 1 - \max_{n_c} \text{silhouette}(\text{KMeans}(n_c))\]

where $w_1 = 0.20$, $w_2 = 0.25$, $w_3 = 0.10$, $w_4 = 0.20$, $w_5 = 0.10$, $w_6 = 0.05$, $w_7 = 0.05$, and $w_8 = 0.05$ are the learned weights.

\end{enumerate}


\subsection{Performance Heatmap}

\begin{figure}[b]  
\centering
\includegraphics[width=\textwidth,keepaspectratio]{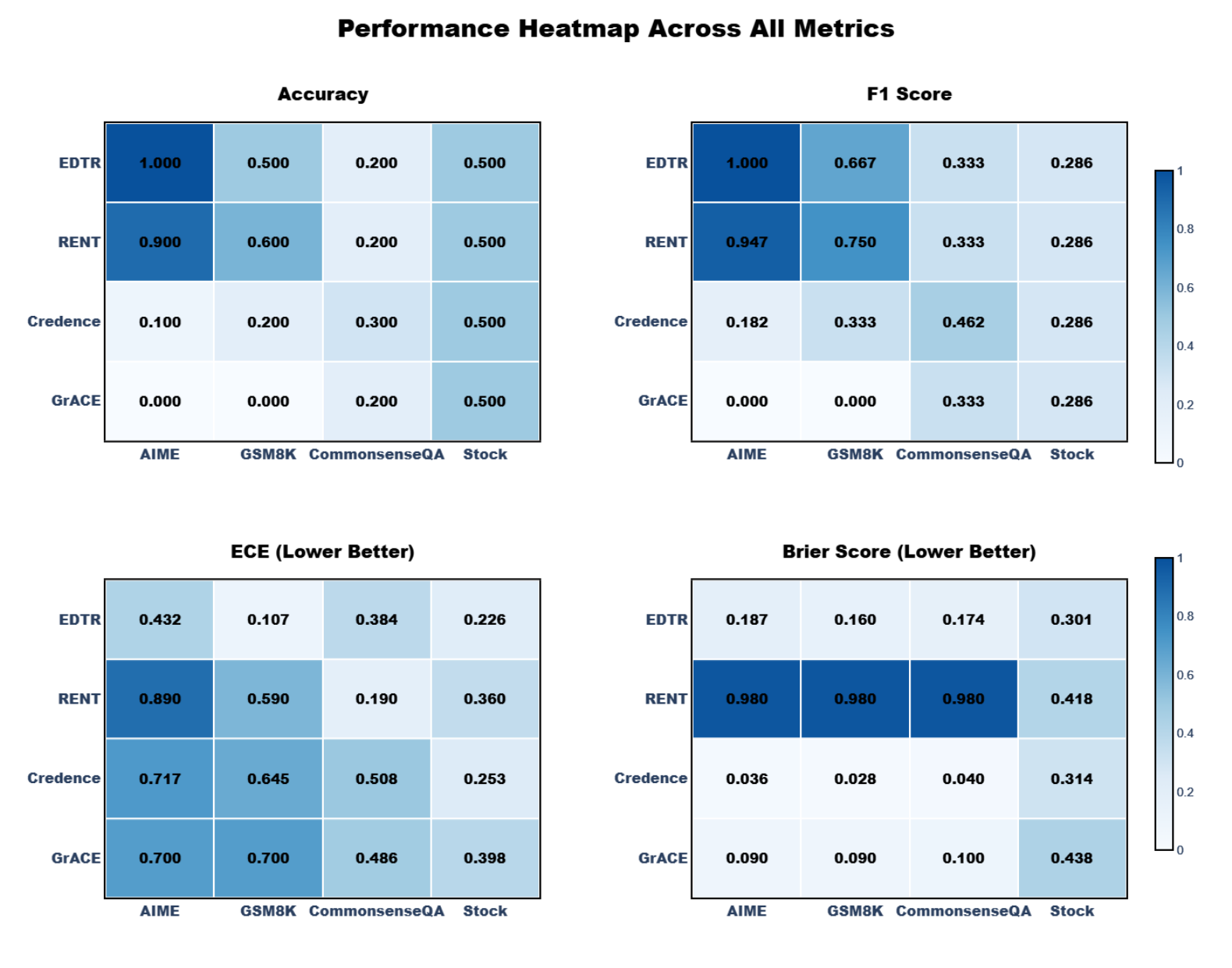}
\caption{Heatmap showing the results for 4 different
metrics. The results are provided across each method
for the four different datasets. The darker the blue,
the closer it is to the ideal result.}
\label{fig:heatmap}
\end{figure}

\begin{ack}
Use unnumbered first level headings for the acknowledgments. All acknowledgments
go at the end of the paper before the list of references. Moreover, you are required to declare
funding (financial activities supporting the submitted work) and competing interests (related financial activities outside the submitted work).
More information about this disclosure can be found at: \url{https://neurips.cc/Conferences/2025/PaperInformation/FundingDisclosure}.

Do {\bf not} include this section in the anonymized submission, only in the final paper. You can use the \texttt{ack} environment provided in the style file to automatically hide this section in the anonymized submission.
\end{ack}






\appendix

\end{document}